\documentclass[letterpaper, 10 pt, conference]{ieeeconf}

\IEEEoverridecommandlockouts

\usepackage{graphics}
\usepackage[normalem]{ulem}
\usepackage{graphicx}
\usepackage{amsmath,amssymb,amsfonts}
\usepackage[caption=false]{subfig}
\usepackage{xcolor}
\usepackage{cite}

\title{\LARGE \bf GazeSplat: Binding Instantaneous Gaze to Persistent Object Instances for Out-of-View Mobile Manipulation}

\author{Changyu Liu$^{\dagger}$, Qijin Song$^{\dagger}$, Jingya Wang, and Weibang Bai$^{\ast}$
\thanks{This work was supported by the Key Laboratory of Intelligent Perception and Human-Machine Collaboration (ShanghaiTech University), Ministry of Education. Corresponding author: Weibang Bai {\itshape (E-mail: wbbai@shanghaitech.edu.cn).}}
\thanks{$^{\dagger}$Changyu Liu and Qijin Song contributed equally to this work. 
}
\thanks{All authors are with the School of Information Science and Technology, ShanghaiTech University,  Shanghai 201210, China. }
}

\begin{document}

\maketitle
\raggedbottom
\thispagestyle{empty}
\pagestyle{empty}

\begin{abstract}
Grounding ambiguous references such as ``bring me this one'' is challenging when similar objects are present and the target lies outside the robot's view. We present GazeSplat, a gaze--speech mobile manipulation framework that binds user gaze to persistent object instances in a shared metric 3D Gaussian Splatting map. The proposed gaze cone, with its width informed by eye-tracker calibration, aggregates angularly weighted rendered surface evidence to select instances and reject weakly supported bindings. Word-level temporal alignment associates these bindings with spoken references. The resulting bindings persist throughout navigation, target re-observation, and manipulation. In scenes containing visually similar objects without identification
markers, cone voting achieves 85.2\% accuracy over 330 stationary
trials and 98.2\% over 57 walking trials, outperforming five grounding baselines,
with the largest gains at minimum angular
separations below $1^\circ$. On a quadruped mobile manipulator, GazeSplat completes 36 of 38 tasks across gaze--speech and named requests, including 15 of 16 gaze--speech tasks initiated with the robot outside the task room.
\end{abstract}

\section{Introduction}

Object retrieval and delivery are fundamental capabilities for mobile manipulation, but grounding the user's reference in a specific object remains challenging when the robot and user do not share a viewpoint. When multiple objects are present, an instruction such as ``Bring me this one'' is ambiguous to a robot without access to the user's gaze. This problem becomes especially difficult in out-of-view manipulation scenarios, where the robot may be operating elsewhere and cannot reliably ground the user's reference using its current visual observations. The key challenge is to associate the user's instantaneous gaze with the corresponding object instance in a persistent, viewpoint-independent 3D representation, and to preserve this association until the robot can navigate to and manipulate the target. Human gaze provides a natural cue for identifying an intended object, often preceding pointing, reaching, or verbal reference~\cite{land1999roles,hayhoe2005eye}. Wearable eye trackers capture this cue continuously and unobtrusively during natural interaction~\cite{kassner2014pupil}.

Gaze information has been employed as a cue for object referencing
and manipulation in human--robot interaction~\cite{belardinelli2024,fischerjanzen2024,famhri,gaze2act2025, glancesay}. 
Visual, semantic and instance-level maps 
support the localization, avoidance, and manipulation of previously observed objects~\cite{wang2024prior,okrobot2024,dynamem2024, song2025map, dovsg}.
These capabilities are complementary: gaze conveys the user's
immediate object reference, while persistent maps retain object
identities and locations for later action. However, a stored map
alone does not determine which instance the user currently intends.
The challenge is to connect real-time gaze to a specific mapped
instance and preserve the resulting reference throughout
subsequent robot execution.\looseness=-1

\begin{figure}[!t]
    \centering
    \includegraphics[width=0.9\columnwidth]{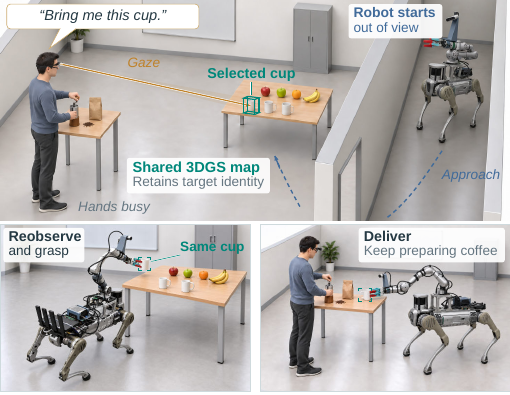}
    \vspace{-3mm}
    \caption{GazeSplat for out-of-view mobile manipulation.
    The request ``Bring me this cup'' is grounded through gaze
    in a persistent object instance within a shared 3DGS map,
    enabling the robot to navigate to and retrieve the same cup
    despite lacking an initial view of the target.}
    \label{fig:teaser}
    \vspace{-5mm}
\end{figure}

To address this gap, we present GazeSplat, a gaze--speech mobile manipulation framework that binds the user's instantaneous gaze to persistent object instances in a shared metric 3D Gaussian Splatting (3DGS) map. The bound instance provides a common reference for speech interpretation and robot execution. Word-level temporal alignment associates gaze-selected instances with spoken references, while stored instance positions guide navigation and target selection during re-observation. This reference remains actionable as the robot's viewpoint and location change, supporting tasks in which the target is outside its view at instruction time.

Reliable instance binding is challenging when candidate objects have small angular separations from the user's viewpoint. Gaze error can cause a single ray to miss the intended surface or intersect a competing instance, leaving target selection ambiguous even when instance identities are known. We introduce a cone-based gaze-to-instance resolution method that aggregates angularly weighted surface evidence rendered from the user's viewpoint. Eye-tracker calibration informs the query width to accommodate directional error, while normalized instance scores and vote shares support candidate selection and rejection of insufficiently supported bindings.

Our main contributions are threefold:

\begin{itemize}
    \item We develop GazeSplat, an end-to-end gaze–speech mobile manipulation framework that binds the user's gaze to persistent object instances in a metric 3DGS map shared with the robot. This framework connects gaze-based object reference to navigation, target re-observation, and manipulation, allowing the robot to act on the referenced instance across viewpoints and rooms.

    \item We introduce a cone-based gaze-to-instance resolution method that aggregates angularly weighted rendered surface evidence from the user's viewpoint. A query width informed by eye-tracker calibration accommodates directional error, while normalized instance scores and relative vote shares support candidate ranking and rejection of insufficiently supported bindings.

    \item We evaluate GazeSplat through gaze-to-instance grounding and real-robot task execution. Cone voting outperforms five alternative grounding strategies, achieving 85.2\% accuracy over 330 stationary trials and 98.2\% over 57 walking trials. The complete system successfully executes 36 of 38 tasks, including tasks with targets initially outside the robot's view, demonstrating that persistent gaze-grounded instance references can support navigation, target re-observation, and manipulation.

\end{itemize}

\section{Related Work}

\subsection{Gaze-Guided Object Reference and Manipulation}

Human gaze provides spatial cues for resolving object references
in robot interaction. FAM-HRI and Glance-Say combine gaze-based
target selection with spoken action instructions~\cite{famhri,glancesay}, while GAMMA uses gaze and a
vision--language model to select and parameterize manipulation
skills~\cite{gamma}. Gaze has also been incorporated into robot
policy learning: GazeVLA represents human intent through
gaze~\cite{gazevla}, and Gaze2Act conditions a
vision--language--action policy on first-person gaze through
cross-view matching~\cite{gaze2act2025}. For mobile object
retrieval, iBotAssistant combines gaze and speech with visual
markers attached to candidate objects~\cite{ibotassistant}. In contrast to
gaze-conditioned policy learning, GazeSplat resolves gaze into an
explicit reference to a persistent map instance, which guides
subsequent navigation and target re-observation before manipulation.\looseness=-1

\subsection{Persistent Spatial Memory for Mobile Manipulation}

Persistent spatial memory provides object information beyond
the robot's current field of view. OK-Robot uses a pre-scanned
semantic voxel map for open-vocabulary pick-and-place
tasks~\cite{okrobot2024}; DynaMem maintains a spatio-temporal
semantic memory that can be updated as the environment
changes~\cite{dynamem2024}; and DovSG represents object instances
and their spatial relationships using a dynamic 3D scene
graph~\cite{dovsg}. For 3DGS-based spatial representations,
LEGS embeds language features into a room-scale Gaussian map
for open-vocabulary object localization~\cite{legs}, while
HAMMER fuses observations from heterogeneous robots and visual
devices into a shared metric-semantic 3DGS map~\cite{hammer}.
These representations retain semantic and spatial information
that can be queried beyond the robot's current observation.

Existing approaches provide different mechanisms for specifying
targets in spatial memory. FARM retrieves previously observed
objects through relational spatial memory~\cite{farm};
HoloSpot allows users to manually select instances in a
pre-scanned model~\cite{holospot}; and MIEL combines speech,
skeletal pointing, and clarification dialogue to resolve
references to objects outside the robot's field of view~\cite{miel}. Compared with manual selection and pointing-based interaction,
GazeSplat combines gaze and speech to bind user references to
specific object instances in a shared map, retaining these
instance references to support subsequent navigation, target
re-observation, and manipulation.

\subsection{World-Referenced Gaze and Object Reference}

Previous studies have investigated how wearable gaze measurements
can be mapped into three-dimensional environments. Wang et al.
used visual--inertial SLAM to estimate the eye-tracker pose and
projected gaze rays into an \mbox{RGB-D} point cloud to obtain
world-frame gaze locations~\cite{wang3dgaze}. 4D Attention
combined wearer localization, scene rendering, and dynamic
object reconstruction to associate gaze with static and
dynamic objects~\cite{four_d_attention}. The Aria Everyday
Activities dataset provides globally aligned wearer trajectories,
3D gaze, timestamped speech, and scene point clouds across
repeated recordings of the same locations~\cite{aria_everyday}.
These studies provide a spatial basis for relating the user's
gaze to the surrounding environment and its objects.

For gaze-guided interaction, measurement uncertainty introduces ambiguity when multiple candidate objects lie near the gaze direction. Glance-Say stabilizes target selection by accumulating distance and directional evidence in the current egocentric image~\cite{glancesay}, whereas BayesGaze accumulates gaze posteriors over predefined two-dimensional targets
~\cite{bayesgaze}. GazeSplat renders the shared 3D map from the wearer's viewpoint and uses angularly weighted surface voting to determine which persistent object instance is indicated by gaze.

\section{Method}
\begin{figure*}[t]
    \centering
    \includegraphics[width=0.85\textwidth]{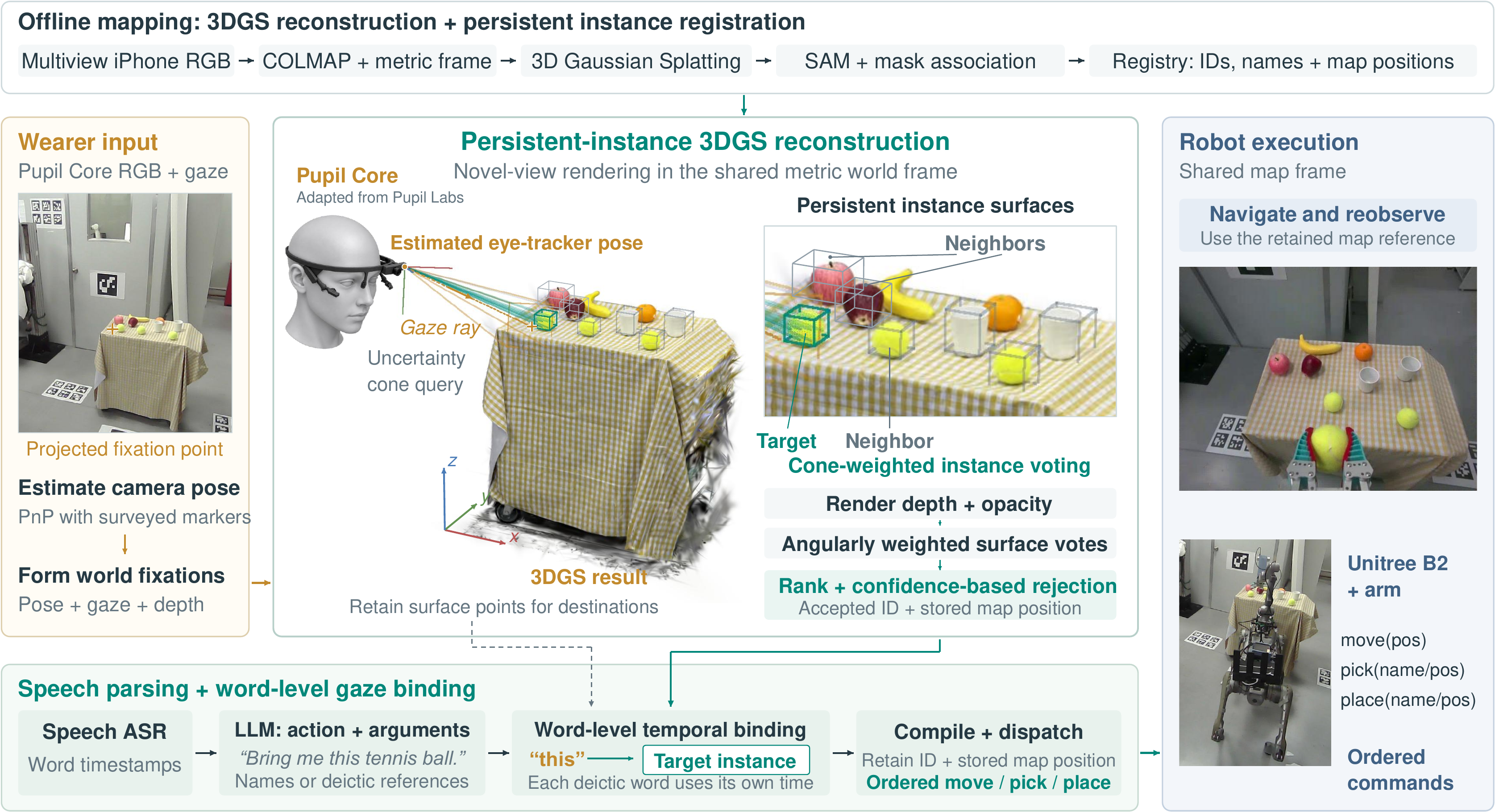}
    \vspace{-2mm}
    \caption{Overview of GazeSplat. Offline mapping constructs a
    persistent-instance 3DGS map in a shared metric frame.
    Online cone-based gaze resolution and word-level temporal
    alignment bind spoken references to mapped object instances.
    The resolved instance identity is retained throughout execution,
    and its stored map position guides robot navigation and target
    re-observation before manipulation.}
    \label{fig:overview}
    \vspace{-6mm}
\end{figure*}

\subsection{Problem Formulation and System Overview}
\label{sec:system_overview}

We consider gaze--speech-driven mobile manipulation in a
pre-mapped environment, with the eye tracker and robot localized
in a shared metric 3DGS map frame. The user selects an object
instance through gaze and specifies an action through speech,
potentially while the target is outside the robot's view.
Inputs comprise timestamped first-person images, gaze measurements,
and spoken instructions with word-level timestamps. The output
is an ordered sequence of move, pick, or place commands with
semantic or spatial arguments.\looseness=-1

As illustrated in Fig.~\ref{fig:overview}, the offline stage
constructs the map and registers object and place instances
(Sec.~\ref{sec:mapping}). Online processing aggregates gaze-event
evidence through cone queries to accept instance bindings
(Sec.~\ref{sec:gaze_resolution}), which are associated with spoken
references through word-level temporal alignment
(Sec.~\ref{sec:gaze_speech}). The bound instance identity is retained during execution,
while its stored position guides navigation and target selection
during re-observation. Manipulation uses current visual
measurements (Sec.~\ref{sec:robot_execution}).

\subsection{Persistent-Instance 3DGS Map}
\label{sec:mapping}

We adopt 3DGS as the shared scene representation to support
surface queries from the wearer's viewpoint, persistent instance
labels tied to scene geometry, and stored metric positions for
robot execution. Its novel-view rendering
capability~\cite{kerbl2023gaussians} supports queries at specified
camera poses, where a single pixel corresponds to a query ray
and a local patch samples neighboring rays. Existing training
and rendering tools~\cite{tancik2023nerfstudio} facilitate map
construction from multi-view phone images and let us reuse
Gaussian projection, depth ordering, and opacity accumulation
for online queries.

Rendering provides depth and accumulated opacity along sampled
directions, accounting for occlusion through depth ordering
and opacity accumulation. Instance votes are collected from
surface points recovered from this depth, making evidence
collection dependent on the scene rendered from the wearer's
viewpoint. During mapping, SAM masks are associated with
Gaussian primitives through rendered depth; during online
queries, recovered surface points are matched to the nearest
Gaussians to retrieve instance labels. Map annotation and
online lookup thus share the same Gaussian geometry, linking
surface queries to persistent object identities. Accumulated
opacity also modulates sample contributions in the voting
procedure described in Sec.~\ref{sec:gaze_resolution}.

We first calibrate the smartphone camera intrinsics and capture
multi-view images of the environment. COLMAP recovers camera poses
and sparse scene geometry with the focal length and principal point
held fixed~\cite{schoenberger2016sfm}. We define the world origin at
a corner of a ChArUco board, triangulate the board corners across
registered views, and align the reconstruction with the board's
known metric coordinates using an Umeyama similarity
transformation~\cite{umeyama1991least}. Following alignment, we train
the 3DGS map using splatfacto in
Nerfstudio~\cite{tancik2023nerfstudio}. Pose reorientation, scene
recentering, and automatic scale normalization are disabled to
preserve the metric world frame defined by the board.

ArUco corners in the scene are surveyed into the same frame using
the same procedure and subsequently support online eye-tracker
localization and robot frame alignment. ChArUco and ArUco markers
serve only to establish and access the shared coordinate frame.
They are neither attached to task objects nor used to identify
object instances.

Instance identities are established through multi-view mask
association~\cite{kirillov2023sam,gaussiangrouping}. We uniformly
sample views from the mapping images and generate 2D masks using
SAM. Mask pixels are back-projected using rendered depth and
associated with Gaussian primitives. Each mask forms a graph node,
with edges connecting masks from different views when their
Gaussian sets overlap sufficiently. The connected components define
the initial instances. After removing anomalous masks spanning
multiple objects and merging part and whole-object masks, we retain
instances supported by at least three views and satisfying a minimum
size requirement.

Each retained instance $o_i$ receives a unique persistent identifier,
and its Gaussian set $\mathcal G_i$, bounding box, and centroid
$\mathbf c_i$ are recorded. Task-relevant instances are manually
named once after mapping and divided into a target set $\mathcal T$
for object reference and manipulation, and a place set $\mathcal P$
used exclusively to resolve placement destinations, such as
tabletops and boxes. Unnamed instances retain their labels but are
excluded from object-binding outputs. The instance registry is
shared by gaze resolution, speech parsing, and robot execution,
allowing target spatial information to be retrieved through its
persistent identity throughout the task.

\subsection{Cone-Based Gaze-to-Instance Resolution}
\label{sec:gaze_resolution}

\paragraph{Gaze uncertainty and query design}
Directional error can cause a single gaze ray to miss the intended
surface, land on a background region, or intersect a neighboring
object. We use a query cone to include evidence from nearby
directions, with greater weight assigned to directions closer to
the measured gaze axis. This accommodates measurement uncertainty
while retaining the directional information in the gaze estimate.
Because the same angular error produces larger spatial offsets
at greater viewing distances, we define the query neighborhood
and its weights in angular space, using eye-tracker calibration
to inform the kernel width.

\paragraph{World-frame gaze events}
The eye tracker is localized in the shared world frame through PnP
using surveyed ArUco corners. The undistorted gaze direction is then
transformed into a world-frame ray. Each pose update requires at
least one surveyed marker and must pass localization validity checks.
During temporary interruptions, the most recent accepted pose can
be reused for up to $1\,\mathrm{s}$. Samples without a sufficiently
recent pose are discarded. Pose reuse retains the last accepted
estimate without predicting wearer motion during marker loss.

Depth queries along valid gaze rays recover surface samples, which
are grouped into world-frame gaze events. The clustering tolerance
scales with viewing distance to accommodate the spatial displacement
associated with angular measurement errors. For each event, the
camera position of the temporally central sample defines the
representative origin $\mathbf o$, and the mean clustered surface
point $\bar{\mathbf x}$ defines the representative gaze point.
The central query direction and viewing distance are:
\begin{equation}
    \hat{\mathbf d}
    =
    \frac{\bar{\mathbf x}-\mathbf o}
    {\|\bar{\mathbf x}-\mathbf o\|},
    \qquad
    D=\|\bar{\mathbf x}-\mathbf o\|.
    \label{eq:event_direction}
\end{equation}

\paragraph{Angular weighting and query extent}
For each sampled unit direction $\mathbf d_j$, indexed by $j$,
we define $\theta_j$ as its angular deviation from the central
gaze direction $\hat{\mathbf d}$ and $w_j$ as its unnormalized
Gaussian angular weight:
\begin{equation}
    \begin{aligned}
        \theta_j
        &=
        \arccos\left(
            \mathbf d_j^{\mathsf T}\hat{\mathbf d}
        \right),\\
        w_j
        &=
        \exp\left(
            -\frac{\theta_j^2}{2\sigma^2}
        \right).
    \end{aligned}
    \label{eq:angular_weights}
\end{equation}

The kernel width $\sigma$ controls angular weight decay. We use
a fixed $\sigma=1^\circ$ across gaze events in the main configuration,
informed by the approximately $1^\circ$ gaze error observed during
wearer calibration. The query patch has horizontal and vertical
half-fields of view of $2\sigma$. Its default resolution of
$33\times33$ provides an angular sampling interval of approximately
$0.12^\circ$. Wider queries use higher resolutions to maintain
an interval of approximately $0.18^\circ$ or finer.

\paragraph{Surface queries and instance voting}
Local depth and accumulated opacity are rendered from $\mathbf o$,
with $\hat{\mathbf d}$ as the central direction. Each sampled
direction recovers a world-frame surface point $\mathbf x_j$,
which is associated with a map entity through the label of
its nearest Gaussian:
\begin{equation}
    \begin{aligned}
        g_j^*
        &=
        \arg\min_{g\in\mathcal G}
        \|\mathbf x_j-\boldsymbol\mu_g\|_2,\\
        \ell_j
        &=
        L(g_j^*),
    \end{aligned}
    \label{eq:surface_assignment}
\end{equation}
where $\mathcal G$ contains all map Gaussians,
$\boldsymbol\mu_g$ is the center of Gaussian $g$, and $L(g)$
returns its entity label. The assignment $\ell_j=k$ indicates
that sample $j$ belongs to instance $o_k$.

Matching against the complete map preserves the assignments of
locations, background regions, and unnamed components. Only surfaces
assigned to instances in the target set $\mathcal T$ contribute
target votes. Segmentation components belonging to the same entity
are merged at the instance level.

Let $\mathcal V$ contain sample indices passing depth-validity,
nearest-neighbor-distance, and depth-margin checks, and let
$\alpha_j$ denote the rendered accumulated opacity for sample $j$.
Using the indicator function $\mathbb I[\cdot]$, the weighted
vote mass for target instance $o_k\in\mathcal T$ is:
\begin{equation}
    m_k
    =
    \sum_{j\in\mathcal V}
    w_j\alpha_j
    \mathbb I[\ell_j=k].
    \label{eq:instance_mass}
\end{equation}

\paragraph{Scale normalization and binding acceptance}
At the same viewing distance, larger objects generally cover more
query directions and can receive greater raw vote mass. To adjust
for object scale and viewing distance, we first express instance
evidence relative to the total kernel weight of the query patch,
then normalize it by a reference coverage:
\begin{equation}
    W=\sum_j w_j,
    \qquad
    q_k=\frac{m_k}{W},
    \qquad
    S_k=\frac{q_k}{C_k},
    \label{eq:capture_score}
\end{equation}
where $W$ is the total angular weight over all sampled directions
in the query patch, including those that do not yield valid
target votes. The quantity $q_k$ is the weighted vote mass
$m_k$ of instance $o_k$ normalized by $W$.
The representative object radius $r_k$ is a physical length
expressed in the same units as the viewing distance $D$;
together, they determine the reference angular radius
$\arctan(r_k/D)$.

We construct a gaze-centered reference disk with this angular
radius and define $C_k$ as the fraction of total kernel weight
covered by the disk on the same discrete samples, i.e.,
$C_k = W^{-1}\sum_j w_j
\mathbb I[\theta_j \leq \arctan(r_k/D)]$.
The normalized score $S_k=q_k/C_k$ compares the observed
instance evidence with this reference coverage and is used
to rank candidate instances. This normalization adjusts for
the effects of object scale and viewing distance on raw votes.
The estimation of $r_k$ is specified in
Sec.~\ref{sec:experimental_setup}.

Candidates are ranked by $S_k$. Online binding requires the
highest-ranked candidate to meet both a minimum score and a
vote-share requirement. The vote share $\rho_k$ of target instance $o_k$ is:
\begin{equation}
    \rho_k
    =
    \frac{m_k}
    {\sum_{o_l\in\mathcal T}m_l}.
    \label{eq:vote_share}
\end{equation}

A sufficiently high share satisfies the vote-share requirement
directly. A lower share must reach a minimum level and provide
a prescribed multiplicative advantage over the second-ranked
candidate's share. The numerical thresholds are given in
Sec.~\ref{sec:experimental_setup}.

Only accepted predictions enter the object-binding buffer,
which stores their persistent instance identities, map centroids,
and timestamps. Gaze surface points are retained independently
for subsequent spatial-destination resolution.

\subsection{Gaze and Speech Binding}
\label{sec:gaze_speech}

A language model parses spoken requests into actions and object
or destination arguments. Parses are cached by transcript, while
stop commands bypass language-model parsing. Explicit names
provide semantic arguments; deictic references are resolved through
word-level gaze alignment. Placement commands reuse the grasped object without resolving ``it'';
destinations use registered places or independently retained gaze
surface points.

Each gaze-dependent reference is resolved independently. The search
window extends from $2\,\mathrm{s}$ before the reference time to
$0.6\,\mathrm{s}$ after it for object and location references,
or $1.5\,\mathrm{s}$ after it for delivery-destination references.
Reference times use word-level timestamps, falling back to the
utterance end time when unavailable. Only records already received
are considered, without waiting for the window to elapse.

For object references, accepted fixations on the same instance are merged into \emph{visits} for gaps up to 0.15\,s, so that a blink does not split a visit.
Eligible visits must overlap the search window, accumulate at
least $0.3\,\mathrm{s}$ of dwell, correspond to a named instance,
and match any specified noun class. We select the temporally
closest eligible visit: distance is zero if it spans the reference
time, and otherwise equals the gap to its nearest boundary.
Ties favor the later visit. Dwell affects eligibility only,
not ranking. Without an eligible visit, the object reference
remains unresolved.

For gaze-based location and delivery-destination references,
we select the surface with the greatest accumulated dwell within
the window and use its last in-window gaze landing point as the
target position. An empty window falls back to the temporally
closest available gaze landing point.

Object and destination arguments may use different gaze records.
Resolved actions and arguments form an ordered command sequence,
with object references retaining instance identities and map
positions for subsequent execution.

\subsection{Robot Execution}
\label{sec:robot_execution}

\begin{figure}[t]
    \centering
    \includegraphics[width=0.95\linewidth]{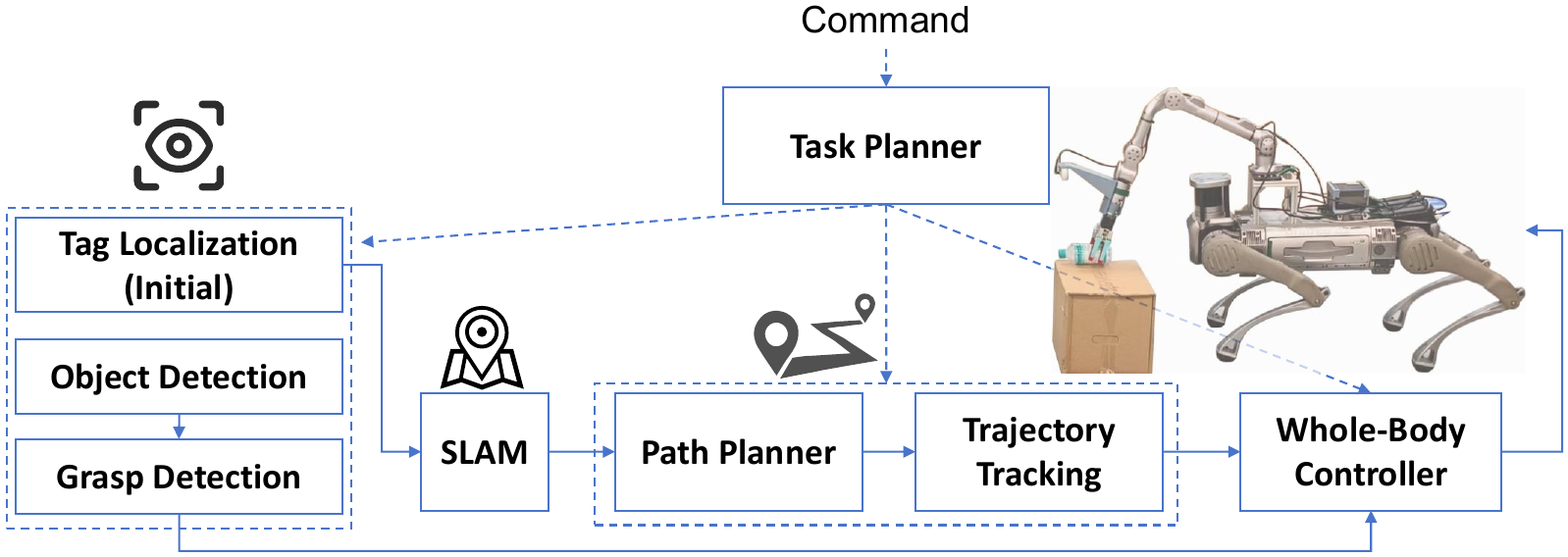}
    \vspace{-2mm}
    \caption{Robot execution using the retained map reference. Task-level commands are dispatched to navigation, perception, and manipulation modules. }
    \label{fig:robot_exe}
    \vspace{-4mm}
\end{figure}

Robot execution uses the spatial reference associated with the resolved instance to connect the user's selection to physical interaction. The framework supports three primitive commands: $\mathtt{move(pos)}$, $\mathtt{pick(name/pos)}$, and $\mathtt{place(name/pos)}$. As shown in Fig.~\ref{fig:robot_exe}, a task planner parses each command, resolves its semantic or spatial arguments, and dispatches it to the appropriate navigation and manipulation modules. For a gaze-grounded object, the stored instance centroid supplies the reference position. A category-name argument allows selection of an available instance of that category, whereas a reference position is used to disambiguate a particular object or specify a target location.

For navigation, the robot pose and the surrounding occupancy map are estimated using LIO-SAM~\cite{lio-sam}. At start-up, the LIO-SAM frame is rigidly aligned with the 3DGS world frame using a PnP pose estimated from the surveyed ArUco markers described in Sec.~\ref{sec:mapping}. Subsequent robot poses are transformed into this shared frame, allowing stored instance centroids and command target positions to serve as spatial references for navigation. Given a target position, an $A^*$-based planner~\cite{astar} computes a collision-free path, which is converted into base velocity commands by a pure-pursuit controller~\cite{macenski2023regulated} using online pose feedback. This navigation pipeline is used both for explicit $\mathtt{move}$ commands and for approaching objects or placement regions outside the manipulator workspace.

Object locations are obtained by combining RGB-based detection with depth measurements. Specifically, a YOLOE detector~\cite{yoloe} predicts object classes and image-space bounding boxes. For each selected detection, valid depth measurements within the bounding box are back-projected using the camera intrinsics, and the resulting 3D position is transformed from the camera frame to the robot/map frame. For name-based picking, an accessible instance of the requested category is selected; when a reference position is provided, the detection spatially closest to that position is used. The estimated object position is then passed to the manipulation module. The retained map reference identifies the intended target during re-observation, and the robot's current RGB-D measurements provide its position for grasp execution.

For manipulation, the task planner decomposes each operation into approach, grasp/release, and retreat motions. The manipulator is controlled by a model predictive controller (MPC)\cite{Song2026quadmpc}, which tracks the desired motion while respecting the robot's kinematic and control constraints. Execution status is returned to the task planner after each primitive, enabling the subsequent instruction to be issued only after the current navigation or manipulation action has completed.

\section{Experiments}
\label{sec:experiments}
\label{sec:experimental_setup}

We evaluate gaze grounding (E1) and robot task execution (E2)
using a Pupil Core eye tracker, wireless microphone, and Unitree
B2 quadruped with a Z1 arm and RealSense camera. Interaction
processing runs on a Linux workstation and execution onboard;
an RTX 4090 workstation supports map training and offline replay.
Six participants provided recordings across independently mapped
tabletop layouts containing identical tennis balls and white cups,
fruit, and other objects, with varying viewing distances and
directions. Task objects carry no identification markers;
environmental markers establish the shared coordinate frame.
Five-point calibration gives an observed gaze error of approximately
$1^\circ$, motivating the default $\sigma=1^\circ$.

For each instance, $r_k$ is estimated as the 84th percentile
of distances from Gaussian centers in its largest segmentation
component to that component's centroid. The largest component
is selected by Gaussian count.
Whisper small transcribes speech locally via faster-whisper on
the CPU with INT8 quantization and beam size 1. Its word-level
timestamps are converted to wall-clock time for gaze alignment.
GPT-5 mini (\texttt{gpt-5-mini}) parses transcripts via the
OpenAI API with \texttt{reasoning\_effort=minimal} and a fixed
JSON schema. The prompt includes supported actions and registered
object names; gaze--speech binding resolves deictic arguments.

E1 evaluates predictions before online acceptance filtering;
E2 uses the complete pipeline. The highest-scoring candidate
must reach 0.2 in normalized score. Its vote share must reach
0.45, or 0.35 with at least 1.4 times the second-ranked
candidate's share. Other predictions do not enter the object-binding buffer.

\subsection{E1: Gaze-to-Instance Grounding}
\label{sec:disambiguation_results}
\label{sec:ablation_results}

\begin{table*}[!t]
    \centering
    \caption{Grounding outcomes by angular separation and motion. Counts: correct / incorrect / no output.}
    \label{tab:query_results}
    \footnotesize
    \setlength{\tabcolsep}{4pt}
    \renewcommand{\arraystretch}{1.15}
    \vspace{-3mm}
    \begin{tabular}{@{}p{0.30\textwidth}ccccc@{}}
        \hline
        \textbf{Method}
        & \shortstack{Stationary\\$\theta<1^\circ$}
        & \shortstack{Stationary\\$1^\circ\leq\theta<2.5^\circ$}
        & \shortstack{Stationary\\$\theta\geq2.5^\circ$}
        & \shortstack{Walking\\$1^\circ\leq\theta<2.5^\circ$}
        & \shortstack{Walking\\$\theta\geq2.5^\circ$} \\
        \hline
        Nearest-centroid matching
        & 19 / 59 / 10
        & 76 / 26 / 1
        & 130 / 9 / 0
        & 32 / 6 / 0
        & 19 / 0 / 0 \\
        Spherical-neighborhood voting
        & 20 / 57 / 11
        & 77 / 23 / 3
        & 104 / 34 / 1
        & 32 / 6 / 0
        & 16 / 3 / 0 \\
        Single-ray query
        & 7 / 19 / 62
        & 29 / 9 / 65
        & 80 / 3 / 56
        & 1 / 1 / 36
        & 12 / 0 / 7 \\
        BayesGaze
        & 46 / 35 / 7
        & 90 / 12 / 1
        & 131 / 4 / 4
        & 36 / 2 / 0
        & 19 / 0 / 0 \\
        Glance-Say
        & 38 / 31 / 19
        & 72 / 12 / 19
        & 125 / 2 / 12
        & 29 / 5 / 4
        & 17 / 1 / 1 \\
        GazeSplat w/o angular weighting
        & 44 / 32 / 12
        & 86 / 15 / 2
        & 132 / 7 / 0
        & 35 / 3 / 0
        & 19 / 0 / 0 \\
        GazeSplat w/o scale normalization
        & 55 / 23 / 10
        & 96 / 6 / 1
        & 132 / 7 / 0
        & 34 / 4 / 0
        & 18 / 1 / 0 \\
        \textbf{GazeSplat full method}
        & \textbf{57 / 21 / 10}
        & \textbf{92 / 10 / 1}
        & \textbf{132 / 7 / 0}
        & \textbf{37 / 1 / 0}
        & \textbf{19 / 0 / 0} \\
        \hline
    \end{tabular}
    \vspace{-2mm}
\end{table*}

The main E1 comparison evaluates instance selection before
acceptance filtering, followed by a separate visit-level analysis
of rejection. Participants followed randomized target-order cards,
including 15-fixation sequences over three tennis balls, each
appearing five times, and mixed-object sequences. Recordings cover
frontal and oblique stationary views and forward, backward, or
sideways walking, yielding 330 stationary and 57 walking trials.

Each trial is scored using the predicted instance with the longest
cumulative duration among the gaze events assigned to that trial.
Outcomes are correct, incorrect, or no output. A trial is correct
when its prediction matches the instructed target; no output
denotes a query that explicitly returns no object instance.
All remaining trials are scored as incorrect. All methods use
the same event-assignment and scoring rules, and all trials
remain in the accuracy denominator.

The minimum angular separation $\theta$ is the angle between
the target-centroid direction and the nearest competing-instance
direction, viewed from the head position. Walking trials span
$1.14^\circ$--$7.54^\circ$, with no trials below $1^\circ$.

We compare GazeSplat with three geometric query baselines and
two gaze-selection baselines, BayesGaze~\cite{bayesgaze} and
Glance-Say~\cite{glancesay}. The geometric baselines use the same
gaze events as GazeSplat. Nearest-centroid matching selects the
named instance closest to the recovered 3D gaze point, subject
to a $0.35\,\mathrm{m}$ distance threshold. Spherical-neighborhood
voting selects the instance with the largest inverse-distance-weighted
vote from labeled Gaussians within $0.20\,\mathrm{m}$ of that point,
without view-dependent visibility filtering. The single-ray query
assigns an instance label to the surface point recovered along the
central gaze direction.
BayesGaze uses $\sigma_{\mathrm{B}}=0.40^\circ$;
Glance-Say is adapted with $1^\circ$ region expansion,
clipped combined evidence, and pixel-distance tie-breaking.
We additionally ablate angular weighting
and scale normalization to assess their individual contributions.

\begin{figure}[!htbp]
    \centering
    \includegraphics[width=0.85\linewidth]{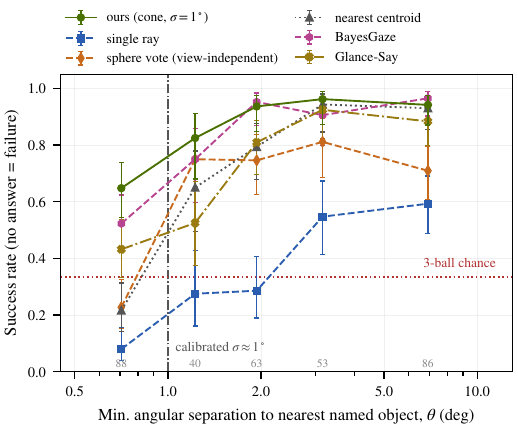}
    \vspace{-3mm}
    \caption{Grounding accuracy versus minimum angular separation $\theta$.
Points are per-bin success rates, error bars are 95\% Wilson
intervals, and grey numbers give the number of trials in each bin.}
    \label{fig:exp_theta}
    \vspace{-3mm}
\end{figure}

\begin{figure}[!htbp]
    \centering
    \includegraphics[width=0.85\linewidth]{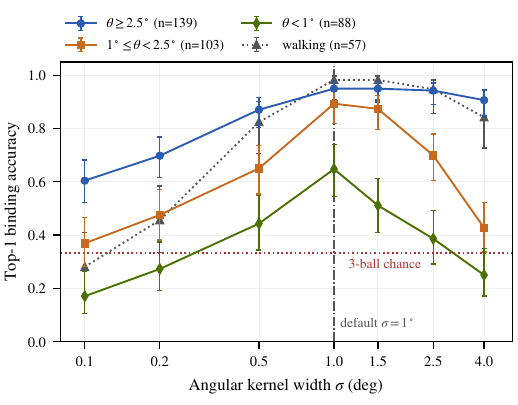}
    \vspace{-4mm}
    \caption{Query-width sensitivity by angular separation
    and motion condition; default $\sigma=1^\circ$.}
    \label{fig:exp_sigma}
    \vspace{-6mm}
\end{figure}

Table~\ref{tab:query_results} shows that GazeSplat achieves 85.2\%
stationary accuracy, compared with 80.9\% for BayesGaze and 71.2\%
for Glance-Say. The advantage is largest below $1^\circ$, where
GazeSplat achieves 64.8\%, versus 52.3\% and 43.2\%, respectively.
The three geometric query baselines achieve \mbox{8.0\%--22.7\%} in this
bin (Fig.~\ref{fig:exp_theta}).

\begin{figure*}[!t]
    \centering
    \includegraphics[width=0.80\textwidth]{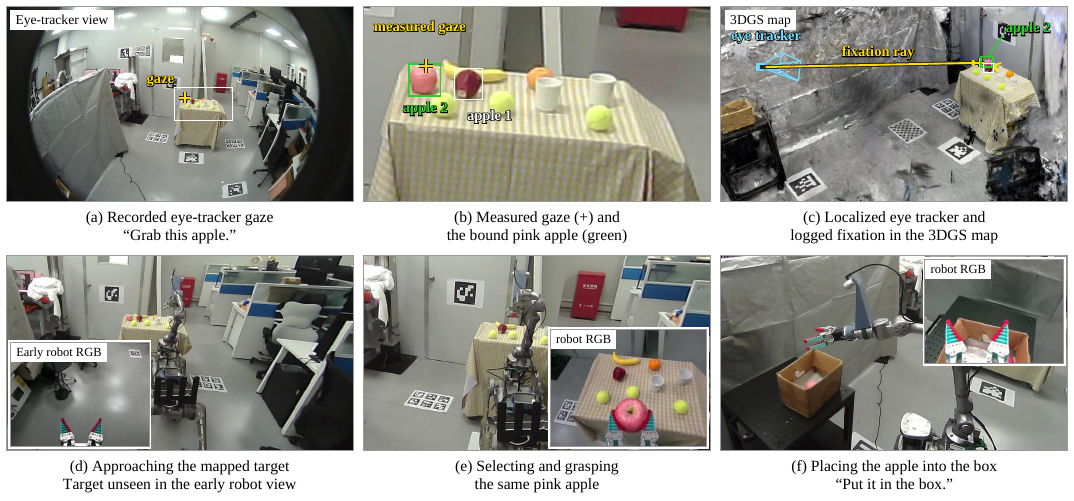}
    \vspace{-3mm}
    \caption{Cross-view retrieval and placement in E2.
    Gaze and speech bind the request to the pink apple in the shared
    3DGS map. The retained map reference guides navigation
    and target selection during re-observation. The robot
    uses current RGB-D measurements to grasp the same apple
    and place it in the requested box. Insets show robot
    views, including an early view with the target outside
    the field of view. Yellow: gaze; cyan: eye tracker;
    green: bound instance.}
    \label{fig:exp_story}
    \label{fig:story}
    \vspace{-3mm}
\end{figure*}

The single-ray baseline primarily fails to return an instance:
across 330 stationary trials, it yields 116 correct, 31 incorrect,
and 183 no-output outcomes, compared with 281 correct,
38 incorrect, and 11 no-output outcomes for GazeSplat.
Directional gaze error can cause a ray to miss target objects
and land on background or unnamed surfaces. Recovered surface
points may also fail the distance checks required for instance
label assignment. Cone voting aggregates surface evidence from
neighboring directions, reducing no-output outcomes and
improving instance selection.

In the densest layouts, predictions can remain concentrated on one ball instead of following target changes, confusing other targets with that instance. Using map-based ray casting, we separate clear and partially occluded views. Below $1^\circ$, the full method achieves 54/70 (77.1\%) accuracy for clear views and 3/18 (16.7\%) for partially occluded views.

Walking and stationary trials use different layouts, so their
results are reported separately. For walking trials, the full
method achieves 97.4\% and 100.0\% accuracy in the two angular
bins. In the $1^\circ\leq\theta<2.5^\circ$ bin,
nearest-centroid matching and spherical-neighborhood voting
each produce six incorrect predictions, compared with one
for GazeSplat. Single-ray querying produces no output in
36 of the 38 trials.

The ablations in Table~\ref{tab:query_results} show that angular
weighting increases stationary accuracy below $1^\circ$ from
50.0\% to 64.8\%, supporting its role in distinguishing closely
spaced instances. Scale normalization increases walking accuracy
from 91.2\% to 98.2\%, but does not consistently improve stationary
performance: overall stationary accuracy is 85.8\% without
normalization and 85.2\% with the full method.

The $\sigma$ sweep (Fig.~\ref{fig:exp_sigma}) shows that narrow
queries miss target surfaces, whereas broader queries admit
more distractors, supporting a query width near the gaze-error
scale.

We additionally assess acceptance filtering on 396 stationary
visits, formed by merging consecutive records for the same instance
using the trial-level event-assignment rule. A trial may contain
multiple visits. Trial-level scoring selects the instance with
the longest cumulative duration, so brief errors may not affect
its outcome. Visit-level evaluation counts these errors individually;
online binding cannot rely on a completed-trial summary.
Filtering reduces incorrect acceptances from 75 to 36 and the
error rate among accepted visits from 18.9\% to 10.6\%
(Table~\ref{tab:acceptance_filter}). It rejects 39 incorrect
and 17 correct visits, retaining 340/396 (85.9\%) overall.

\begin{table}[!htbp]
    \centering
    \vspace{-2mm}
    \caption{Acceptance filtering on 396 stationary visits.
    Correct and Incorrect count accepted visits.
    Error is the fraction of accepted visits that are incorrect.}
    \label{tab:acceptance_filter}
    \small
    \setlength{\tabcolsep}{3pt}
    \renewcommand{\arraystretch}{1.1}
    \begin{tabular}{@{}ccccc@{}}
        \hline
        \textbf{Filter}
        & \textbf{Correct}
        & \textbf{Incorrect}
        & \textbf{Rejected}
        & \textbf{Error} \\
        \hline
        Off & 321 & 75 & 0  & 18.9\% \\
        On  & 304 & 36 & 56 & 10.6\% \\
        \hline
    \end{tabular}
    \vspace{-5mm}
\end{table}

\subsection{E2: Real-Robot Task Execution}
\label{sec:end_to_end_results}

E2 focuses on gaze--speech requests such as \emph{Bring me this tennis ball} and \emph{Put it here}, with named requests such as \emph{Bring me the apple} and \emph{Put it in the box} additionally testing spatial-memory lookup and execution. The robot starts either inside or outside the task room, and results are grouped by its actual starting location. The accompanying video shows representative tasks of each type. Each task is attempted only once, without internal retries or human intervention. Success requires correct request interpretation and grounding, followed by successful completion of all requested navigation and manipulation actions. Any error at any stage, including a parsing error or a failed grasp, causes the entire trial to be marked as failed. Incorrect initial grounding terminates the trial immediately.

\begin{table}[!htbp]
    \centering
    \vspace{-2mm}
    \caption{Task outcomes by reference type and actual
    starting location.}
    \label{tab:e2_results}
    \small
    \setlength{\tabcolsep}{3pt}
    \renewcommand{\arraystretch}{1.1}
    \begin{tabular}{@{}llccc@{}}
        \hline
        \textbf{Reference}
        & \textbf{Start}
        & $n$
        & \shortstack{\textbf{Correct}\\\textbf{target}}
        & \textbf{Completed} \\
        \hline
        \textbf{Gaze--speech}
        & \textbf{Outside}
        & \textbf{16}
        & \textbf{15}
        & \textbf{15} \\
        Gaze--speech
        & Inside
        & 8
        & 7
        & 7 \\
        Named
        & Outside
        & 8
        & 8
        & 8 \\
        Named
        & Inside
        & 6
        & 6
        & 6 \\
        \hline
        \textbf{Total}
        & ---
        & \textbf{38}
        & \textbf{36}
        & \textbf{36} \\
        \hline
    \end{tabular}
    \vspace{-3mm}
\end{table}

Gaze--speech tasks achieve 22/24 completions (91.7\%), including
15/16 (93.8\%) from outside the room
(Table~\ref{tab:e2_results}). Both failures occur during gaze
grounding, one per starting-location group; all correctly
grounded tasks complete execution. Named requests achieve
14/14, giving 36/38 overall (94.7\%).
Figure~\ref{fig:exp_story} illustrates cross-view grasping
and placement of the gaze-selected apple.

Processing-stage medians sum to 2.2 s for cache hits and 5.0 s for cache misses. All 52 stop requests were accepted. Median latencies were 2.0 s from utterance end to stop-command dispatch and 0.13 s from dispatch to the current action's logged transition to a terminal state.

\section{Conclusion}

We presented GazeSplat, a gaze–speech framework for out-of-view mobile manipulation. Cone-based voting and word-level temporal alignment bind user references to instances in a shared metric 3DGS map, preserving target identity throughout navigation, re-observation, and manipulation within a task. The proposed cone voting method outperformed five evaluated baselines, particularly at small angular separations. The system completed 36 of 38 real-robot tasks, including 15 of 16 gaze–speech tasks with the quadruped mobile manipulator starting outside the task room.

The system requires pre-mapped environments and registered instances. It does not update its map for new or moved objects in real time, limiting subsequent manipulation of moved objects. Wearer localization also relies on environmental ArUco markers. Future work will address online map updates, instance re-identification under scene changes, and localization without environmental markers to support more dynamic environments.


\begin{thebibliography}{99}

\bibitem{land1999roles} M. Land, N. Mennie, and J. Rusted, ``The roles of vision and eye movements in the control of activities of daily living,'' \emph{Perception}, vol. 28, no. 11, pp. 1311--1328, 1999.

\bibitem{hayhoe2005eye} M. Hayhoe and D. Ballard, ``Eye movements in natural behavior,'' \emph{Trends Cogn. Sci.}, vol. 9, no. 4, pp. 188--194, 2005.

\bibitem{kassner2014pupil} M. Kassner \emph{et al.}, ``Pupil: An open source platform for pervasive eye tracking and mobile gaze-based interaction,'' in \emph{Proc. ACM Int. Joint Conf. Pervasive Ubiquitous Comput.}, 2014, pp. 1151--1160.

\bibitem{belardinelli2024} A. Belardinelli, ``Gaze-based intention
estimation: Principles, methodologies, and applications in HRI,''
\emph{ACM Trans. Hum.-Robot Interact.}, vol. 13, no. 3, Art. no. 30,
2024.

\bibitem{fischerjanzen2024} A. Fischer-Janzen, T. M. Wendt, and
K. Van Laerhoven, ``A scoping review of gaze and eye tracking-based
control methods for assistive robotic arms,'' \emph{Front. Robot. AI},
vol. 11, p. 1326670, 2024.

\bibitem{famhri} Y. Lai \emph{et al.}, ``FAM-HRI: Foundation-model assisted multi-modal
human-robot interaction combining gaze and speech,'' \emph{IEEE Trans.
Autom. Sci. Eng.}, vol. 23, pp. 10521--10534, 2026.

\bibitem{gaze2act2025} K. Zuo \emph{et al.}, ``Gaze2Act:
Gaze-conditioned vision-language-action policies for interactive robot
manipulation,'' arXiv:2605.30282, 2026.

\bibitem{glancesay} Y. Lai \emph{et al.}, ``Glance-Say: Multimodal human-robot collaboration and intent
recognition via sticky glance,'' arXiv:2603.06121, 2026.

\bibitem{wang2024prior} Y. Wang, W. Bai, Z. Zhang, H. Wang, H. Sun, and Q. Cao, ``Prior-SLAM: Enabling visual SLAM for loop closure under large viewpoint variations,'' \emph{IEEE Trans. Robot.}, vol. 41, pp. 687--707, 2024.

\bibitem{okrobot2024} P. Liu \emph{et al.}, ``OK-Robot: What really matters in integrating
open-knowledge models for robotics,'' in \emph{Proc. Robotics: Science
and Systems (RSS)}, 2024.

\bibitem{dynamem2024} P. Liu \emph{et al.}, ``DynaMem: Online
dynamic spatio-semantic memory for open world mobile manipulation,''
in \emph{Proc. IEEE ICRA}, 2025, pp. 13346--13355.

\bibitem{song2025map} Q. Song and W. Bai, ``Map imagination like blind humans: Group diffusion model for robotic map generation,'' in \emph{Proc. IEEE ROBIO}, 2025, pp. 290--296.

\bibitem{dovsg} Z. Yan \emph{et al.}, ``Dynamic open-vocabulary 3D
scene graphs for long-term language-guided mobile manipulation,''
\emph{IEEE Robot. Autom. Lett.}, vol. 10, no. 5, pp. 4252--4259,
2025.

\bibitem{gamma} T. Y. H. Tay \emph{et al.}, ``Intent at a glance: Gaze-guided robotic
manipulation via foundation models,'' arXiv:2601.05336, 2026.

\bibitem{gazevla} C. Li \emph{et al.}, ``GazeVLA: Learning human intention for robotic
manipulation,'' arXiv:2604.22615, 2026.

\bibitem{ibotassistant} X. Liang and J. Cai, ``A multimodal agentic AI
framework for intuitive human--robot collaboration,'' \emph{Sensors},
vol. 26, no. 6, p. 1958, 2026.

\bibitem{legs} J. Yu \emph{et al.}, ``Language-embedded Gaussian
splats (LEGS): Incrementally building room-scale representations with
a mobile robot,'' in \emph{Proc. IEEE/RSJ IROS}, 2024,
pp. 13326--13332.

\bibitem{hammer} J. Yu, T. Chen, and M. Schwager, ``HAMMER:
Heterogeneous, multi-robot semantic Gaussian splatting,'' \emph{IEEE
Robot. Autom. Lett.}, vol. 10, no. 7, pp. 7270--7277, 2025.

\bibitem{farm} S. He \emph{et al.}, ``FARM: Find anything using
relational spatial memory,'' arXiv:2606.15476, 2026.

\bibitem{holospot} P. Soler Garcia \emph{et al.}, ``HoloSpot:
Intuitive object manipulation via mixed reality drag-and-drop,''
arXiv:2410.11110, 2024.

\bibitem{miel} A. Oyama \emph{et al.}, ``Take that for me: Multimodal exophora resolution with
interactive questioning for ambiguous out-of-view instructions,'' in
\emph{Proc. IEEE RO-MAN}, 2025, pp. 563--570.

\bibitem{wang3dgaze} H. Wang \emph{et al.}, ``SLAM-based localization of 3D gaze using a mobile eye tracker,'' in
\emph{Proc. ACM ETRA}, 2018, Art. no. 65.

\bibitem{four_d_attention} S. Oishi, K. Koide, M. Yokozuka, and
A. Banno, ``4D Attention: Comprehensive framework for spatio-temporal
gaze mapping,'' \emph{IEEE Robot. Autom. Lett.}, vol. 6, no. 4,
pp. 7240--7247, 2021.

\bibitem{aria_everyday} Z. Lv \emph{et al.}, ``Aria Everyday
Activities dataset,'' arXiv:2402.13349, 2024.

\bibitem{bayesgaze} Z. Li \emph{et al.}, ``BayesGaze: A Bayesian
approach to eye-gaze based target selection,'' in \emph{Proc. Graphics
Interface}, 2021, pp. 231--240.

\bibitem{kerbl2023gaussians} B. Kerbl, G. Kopanas, T. Leimk\"uhler,
and G. Drettakis, ``3D Gaussian splatting for real-time radiance field
rendering,'' \emph{ACM Trans. Graph.}, vol. 42, no. 4, Art. no. 139,
2023.

\bibitem{tancik2023nerfstudio} M. Tancik \emph{et al.}, ``Nerfstudio:
A modular framework for neural radiance field development,'' in
\emph{ACM SIGGRAPH Conf. Proc.}, 2023, Art. no. 72.

\bibitem{schoenberger2016sfm} J. L. Sch\"onberger and J.-M. Frahm,
``Structure-from-motion revisited,'' in \emph{Proc. IEEE CVPR}, 2016,
pp. 4104--4113.

\bibitem{umeyama1991least} S. Umeyama, ``Least-squares estimation of
transformation parameters between two point patterns,'' \emph{IEEE
Trans. Pattern Anal. Mach. Intell.}, vol. 13, no. 4, pp. 376--380,
1991.

\bibitem{kirillov2023sam} A. Kirillov \emph{et al.}, ``Segment
anything,'' in \emph{Proc. IEEE/CVF ICCV}, 2023, pp. 4015--4026.

\bibitem{gaussiangrouping} M. Ye, M. Danelljan, F. Yu, and L. Ke,
``Gaussian Grouping: Segment and edit anything in 3D scenes,'' in
\emph{Proc. ECCV}, 2024, pp. 162--179.

\bibitem{lio-sam} T. Shan \emph{et al.}, ``LIO-SAM: Tightly-coupled
lidar inertial odometry via smoothing and mapping,'' in \emph{Proc.
IEEE/RSJ IROS}, 2020, pp. 5135--5142.

\bibitem{astar} P. E. Hart, N. J. Nilsson, and B. Raphael, ``A formal
basis for the heuristic determination of minimum cost paths,''
\emph{IEEE Trans. Syst. Sci. Cybern.}, vol. 4, no. 2, pp. 100--107,
1968.

\bibitem{macenski2023regulated} S. Macenski \emph{et al.},
``Regulated pure pursuit for robot path tracking,'' \emph{Auton.
Robots}, vol. 47, no. 6, pp. 685--694, 2023.

\bibitem{yoloe} A. Wang \emph{et al.}, ``YOLOE: Real-time seeing anything,'' in \emph{Proc.
IEEE/CVF ICCV}, 2025, pp. 24591--24602.


\bibitem{Song2026quadmpc} Q. Song, and W. Bai,``MoMa-MO: LLM-Guided Mobile Manipulation Among Movable Obstacles with Attribute-Grounded Obstacle Memory,'' arXiv: 2611.36101, 2026.

\end{thebibliography}
\end{document}